\pdfoutput=1

\documentclass[11pt]{article}

\usepackage{acl}

\usepackage{times}
\usepackage{latexsym}

\usepackage[T1]{fontenc}

\usepackage[utf8]{inputenc}

\usepackage{microtype}

\usepackage{inconsolata}

\usepackage{algorithm}
\usepackage{algorithmic}

\usepackage{amsmath,amssymb,graphicx}
\usepackage{multirow}
\usepackage{enumitem}

\usepackage{subcaption}

\title{Conformer-Based Speech Recognition \\On Extreme Edge-Computing Devices}

\author {
    Mingbin Xu\thanks{Equal contribution.}\textsuperscript{\rm 1},
    Alex Jin$^*$\thanks{left Apple after paper submission.}\textsuperscript{\rm 1},
    Sicheng Wang\textsuperscript{\rm 1},
    Mu Su\textsuperscript{\rm 1},
    Tim Ng\textsuperscript{\rm 1},
    Henry Mason\textsuperscript{\rm 1},\\
    \textbf{Shiyi Han\textsuperscript{\rm 1},
    Zhihong Lei\textsuperscript{\rm 1},
    Yaqiao Deng\textsuperscript{\rm 1},
    Zhen Huang\textsuperscript{\rm 1},
    Mahesh Krishnamoorthy\textsuperscript{\rm 1}} \\
    \textsuperscript{\rm 1}Apple\\
    mingbinxu@apple.com, alexgbjin@gmail.com, \\
    \{sicheng\_wang,mu\_su,tim\_ng,hmason,shan26,zlei,yaqiao\_deng,zhen\_huang,maheshk\}@apple.com
}
    
\begin{document}
\maketitle

\begin{abstract}
With increasingly more powerful compute capabilities and resources in today's devices, traditionally compute-intensive automatic speech recognition (ASR) has been moving from the cloud to devices to better protect user privacy. However, it is still challenging to implement on-device ASR on resource-constrained devices, such as smartphones, smart wearables, and other small home automation devices. In this paper, we propose a series of model architecture adaptions, neural network graph transformations, and numerical optimizations to fit an advanced Conformer based end-to-end streaming ASR system on resource-constrained devices without accuracy degradation. We achieve over 5.26 times faster than realtime (0.19 RTF) speech recognition on small wearables while minimizing energy consumption and achieving state-of-the-art accuracy.
The proposed methods are widely applicable to other transformer-based server-free AI applications. In addition, we provide a complete theory on optimal pre-normalizers that numerically stabilize layer normalization in any $L_p$-$norm$ using any floating point precision.

\end{abstract}

\section{Introduction}
\label{sec:intro}
Conformer-based \cite{DBLP:conf/interspeech/GulatiQCPZYHWZW20} end-to-end (E2E) automatic speech recognition (ASR) \cite{DBLP:conf/interspeech/YaoWWZYYPCXL21,DBLP:conf/interspeech/ZhangWPSY00YP022} with streaming capabilities \cite{DBLP:conf/icassp/HeSPMAZRKWPLBSL19} have made numerous advances recently. This has paved the way for fully neural speech recognition on resource-constrained mobile devices. These systems also have numerous advantages over conventional hybrid-HMM ASR \cite{DBLP:journals/spm/X12a}.

First, the training procedure is simplified; the entire system can be defined in a single deep learning framework such as PyTorch or TensorFlow. 
Second, recent work (e.g. \citealp{DBLP:conf/interspeech/MiaoCZL019,DBLP:conf/icassp/SainathHLNPBCLA20,DBLP:conf/interspeech/LiZMLWPMWHZG20,DBLP:journals/corr/abs-2310-09988,DBLP:conf/asru/LeiXHLHNZPHDS23}) shows E2E ASR systems can provide better Word-Error-Rate (WER) when compared to conventional hybrid ASR systems. Third, with the continued advancement of deep learning applications, special hardware accelerators such as NVIDIA's Graphics Processing Units (GPU), Google's Tensor Processing Units (TPU), and Apple's Neural Engine (ANE) are becoming increasingly popular. 
A fully neural ASR system can best utilize such hardware advancements and operate with high throughput while minimizing energy consumption.



In this paper, we present optimizations to enable fully E2E neural network based ASR system under resource-constrained environments, such as smartphones, wearables, and home automation devices. Operating fully offline saves cloud computing resources while providing stronger user privacy \cite{DBLP:conf/icassp/XuSTAGDZAHDLWJ23} guarantees, as the user's speech does not need to be transmitted outside of the device.

When targeting resource constrained devices, hardware limitations present many challenges. We describe several multidisciplinary solutions we explored, including memory-aware network transformation, model structural adjustment, and numerical optimizations to address inference stability. We specifically focus on our efforts to take advantage of the inference efficiency provided by specialty hardware accelerators. 
We derive a theory to numerically stabilize computation of layer normalization on hardware accelerators. 
This stabilization technique does not require model retraining and is applicable to the computation of any $L_p$-$norm$. 

\section{Prior Work}
\label{sec:format}
Improving the efficiency of the Transformer architecture has seen substantial interest. \citet{dblp:journals/csur/TayDBM23} provides a comprehensive survey primarily concentrating on model architecture improvements. \citet{DBLP:journals/corr/abs-2302-14017} is another noteworthy resource which delves deeper into considerations specific to hardware configurations.
Linear Transformer \cite{DBLP:conf/icml/KatharopoulosV020}  is a key technique, mitigating the computationally expensive softmax function \cite{bridle1989training} within the attention mechanism. Softmax is also susceptible to numeric overflow problems when computing with limited numerical range. \citet{hoffer2018norm, dblp:conf/nips/ZhangS19a} discuss alternative normalization methods other than Batchnorm \cite{ioffe2015batch} and Layernorm \cite{dblp:journals/corr/BaKH16} to improve computational efficiency and numerical stability in low precision environments. 
Principles for optimizing transformers have been described in \citet{neural-engine-transformers} which target Apple hardware, but are generally applicable for similar devices.
 Within the domain of speech recognition, Squeezeformer \cite{DBLP:conf/nips/KimGSLMMMK22} stands as a seminal work focusing on efficiency optimization, particularly with respect to the Conformer architecture. The paper uses depthwise separable convolution subsampling to substantially save computation which is central to MobileNet \cite{dblp:journals/corr/HowardZCKWWAA17}.
 It's worth mentioning that the majority of prior work focuses on improving training efficiency by making modifications to the existing model architecture. As a result, these changes require model retraining to achieve efficiency improvements. In contrast, our research primarily concentrates on post-training, inference-only processes while avoiding model retraining whenever possible.


\section{Backbone Model}
\label{sec:model}

Our backbone model is built upon the Conformer neural architecture \cite{DBLP:conf/interspeech/GulatiQCPZYHWZW20} as shared acoustic encoder while connectionist temporal classification \cite{DBLP:conf/icml/GravesFGS06} (CTC) and Attention-based Encoder Decoder (AED) \cite{DBLP:conf/icassp/ChanJLV16} as dual decoders trained with multitask learning mechanism \cite{caruana1997multitask}.

Similar to prior work (e.g. \citealp{DBLP:conf/interspeech/GulatiQCPZYHWZW20}), we stack transformer \cite{DBLP:conf/nips/VaswaniSPUJGKP17} layers and convolution \cite{DBLP:journals/pieee/LeCunBBH98} layers alternatively to convert speech frames into high-level representation. 
We use a relative sinusoidal positional encoding \cite{DBLP:conf/acl/DaiYYCLS19} into transformer layers.
Since our goal is to stream ASR on edge devices, we adopt the chunk-based attention strategy to better balance accuracy and dependency of future audio frames \cite{DBLP:conf/interspeech/YaoWWZYYPCXL21,DBLP:conf/interspeech/ZhangWPSY00YP022}.

\section{Proposed Optimizations}
\label{sec:typestyle}


\subsection{Depthwise Separable Convolution}


In the original Conformer encoder design \cite{DBLP:conf/interspeech/GulatiQCPZYHWZW20}, the subsampling module at the beginning of the architecture is implemented using two vanilla convolution layers. 
Our profiling shows that vanilla convolution subsampling accounts for 32.8\% of the overall computation and becomes expensive on resource-constrained devices. To alleviate this bottleneck, we used the idea of depthwise separable convolution \cite{dblp:journals/corr/HowardZCKWWAA17,DBLP:conf/cvpr/Chollet17} as a drop-in replacement and reduced this computational bottleneck to 4.0\% whilst maintaining the WER \cite{DBLP:conf/nips/KimGSLMMMK22}, making it particularly well-suited for inference tasks on mobile devices.

While most of the research emphasizes depthwise separable convolution’s (DWS) computational efficiency and small memory footprint, its effect on reducing dynamic range of the outputs needs more study. 
The possible reason could be that DWS reduces the number of multiply-accumulate operations needed for the convolution filters, hence the chance of bigger values.
Low numeric range is of great importance for model deployment on edge devices equipped with hardware accelerators.
Those hardware often operate in low precision (e.g.fp16) to ease the burden of storage and memory and are exposed to overflow. 


\subsection{Memory-aware Graph Execution}
\label{ssec:graph-exec}

\begin{figure*}[t!]
    \begin{subfigure}[b]{0.9\linewidth}
        \centering
        \includegraphics[width=1.0\textwidth]{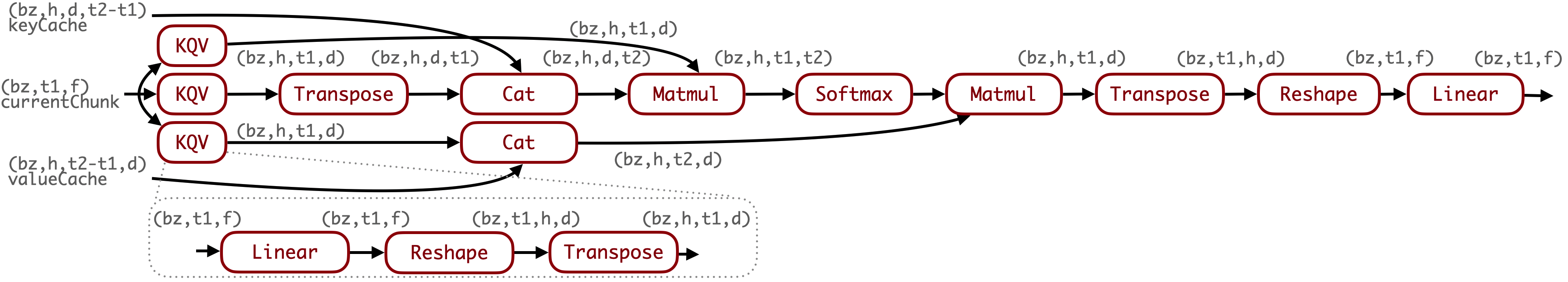}
        \caption{Common compute flow of MHA}
        \label{fig:mha-before}
    \end{subfigure}
    \begin{subfigure}[b]{0.90\linewidth}
        \centering
        \includegraphics[width=1.0\textwidth]{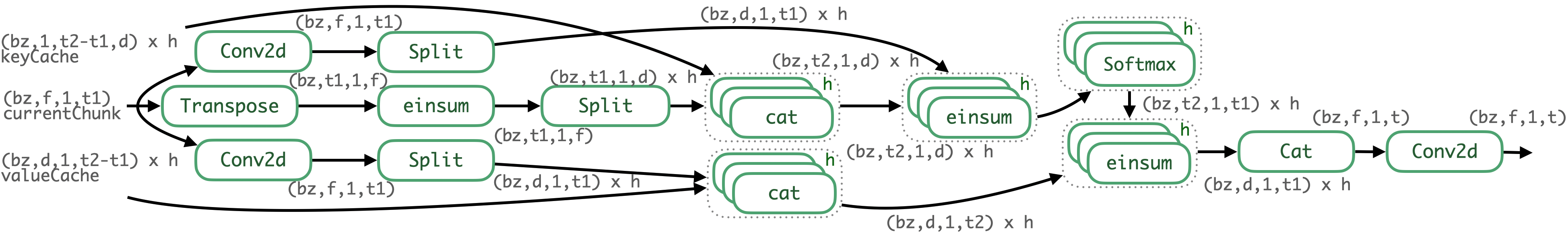}
        \caption{ANE-optimized compute flow of MHA}
        \label{fig:mha-after}
    \end{subfigure}
    \caption{$bz$, $h$ and $f$ refers to batch size, number of attention heads and feature dimension respectively, whereas $d = f / h$. Firstly, we transposed the input and output of Conformer CTC, expanding the input tensor to the desired shape of $(B, C, 1, S)$. This transformation allowed us to execute most layers on the hardware accelerator as per \textit{Principle 1}. Additionally, we extensively employed split and concatenation operations to enhance L2 cache residency (\textit{Principle 2}). To address the issue of undesired memory copies resulting from batched matrix multiplication layers, we replaced them with Einstein summation operations (\textit{Principle 3}).}
    \label{fig:mha}
\end{figure*}

In Apple's white paper \cite{neural-engine-transformers} on deploying transformers on the Apple Neural Engine (ANE), \textit{four principles} are elaborated for optimizing transformers on the ANE:

\begin{itemize}[noitemsep, nolistsep, leftmargin=*]
   \item \textit{Principle 1: Picking the Right Data Format}
   \begin{itemize}
       \item The (B, C, 1, S) \{Batch, Channel, 1, Sequence\} data format is chosen for tensor representation to align with the ANE's 4D and channels-first architecture.
   \end{itemize}
   \item \textit{Principle 2: Chunking Large Intermediate Tensors}
   \begin{itemize}
       \item Utilize split and concatenation operations to divide tensor into smaller chunks and increase L2 cache residency.
   \end{itemize}
   \item \textit{Principle 3: Minimizing Memory Copies}
   \begin{itemize}
       \item Minimize the number of memory operations on tensors such as reshape and transpose.
       \item Represent batch matrix multiplication operations using Einstein summation layers.
   \end{itemize}
   \item \textit{Principle 4: Handling Bandwidth-Boundness}
   \begin{itemize}
       \item We should carefully benchmark the model performance with various batch sizes and sequence lengths and make an informed decision about the cost of memory fetches when we become bandwidth-bound on the ANE.
   \end{itemize}
\end{itemize}
The key idea behind these 4 principles is being aware of high cost invoked by memory copies between CPU and our hardware accelerator. In our implementation, we adhered to the aforementioned principles.
We demonstrate how to rewrite multihead attention (MHA) in Figure \ref{fig:mha} as an example.
More importantly, operations not supported by hardware accelerator were positioned at the beginning or end of the network graph, thus minimizing copies in the memory.

\subsection{Stability of Layer Normalization}
\label{sec:Layernorm}

Layer normalization has become the \textit{de facto} normalization method in transformers after \textit{Attention is all you need} \cite{DBLP:conf/nips/VaswaniSPUJGKP17}. This normalization technique is widely used in the Conformer CTC architecture. On the other hand, modern hardware accelerators for deep learning often exploit lower precision compute paths in order to reduce memory and boost computation throughput. In the Conformer model, we observed that layer normalization and hardware accelerators are often in dissonance with each other. The reason is that skip connections in the Conformer model join values of varying magnitudes to a single tensor and this often leads to numerical \textit{underflows} or \textit{overflows} in low precision compute paths. For example, the maximum value is 65504 in half precision floating point format \cite{4610935}. As a contrast, the maximum value is $3.4e38$ in single precision floating point format.

\begin{align}
\label{eq:Layernorm}
    \hat{x}_i = \frac{x_i-\mu}{\sqrt{\sigma^2+\epsilon}}\ \ \ \ \ \ (Layernorm).
\end{align}

Equation \eqref{eq:Layernorm} is a common realization of layer normalization with respect to the $L_2$-$norm$, where $\mu$ and $\sigma^2$ are the mean and variance of a vector $\mathbf{x}=\{x_i | 1\leq i\leq n, x_i\in \mathbb{R}\}$. A small $\epsilon$ is added at the bottom to avoid division by zero when $\sigma$ is small. In order to compute the variance, however, we need to sum the squares of each $x_i$, which often leads to numerical instability in low precision compute paths. To combat this issue, we employ a technique called Mean Absolute Deviation (MAD) normalization as a pre-normalizer. We note that Layernorm is unaffected by global shifts or global re-scaling of the $x_i$'s and will from here on assume $\mu=0$.

\textbf{Definition 1.} \textit{Given a low precision compute path with a maximum value $M$, an optimal $L_p$-$norm$ pre-normalizer for this compute path maps any distribution of values to a bounded region, $[-D, D]$, where $D$ is as large as possible without causing overflows during the computation of the $L_p$-$norm$.}

We note that in the above definition, we explicitly set a constraint to make $D$ as large as possible to minimize the effect of underflow while staying below our low precision limit.

\textbf{Lemma 1.} \textit{Let $\mathbf{x}=\{x_1, x_2, ..., x_n\}$ be a finite vector of real numbers with $\sum_{i=1}^n x_i=0$, and let $S=\sum_{i=1}^{n} |x_i|$ be its $L_1$-$norm$. Let $p \geq 1$ be a real number. We have}

\begin{align*}
    ||\mathbf{x}||_p^p = \sum_{i=1}^n |x_i|^p \leq 2^{1-p} S^p
\end{align*}

\textit{and the maximum is attained when $\mathbf{x}=\{-\frac{S}{2}, 0, ..., 0, \frac{S}{2}\}$}.

\textit{Proof:} For the cases where $n=1$ or $p=1$, the inequality above trivially holds.

Let's now look at the case where $n\geq 2$ and $p>1$. Let $\mathbf{x}=\{x_1, x_2, ..., x_n\}$ be any vector of real numbers and let $S$ be its $L_1$-$norm$. Consider the vector $\mathbf{v} = \{-\frac{S}{2}, 0, ..., 0, \frac{S}{2}\}$, then

\begin{align*}
    ||\mathbf{v}||_p^p = 2 (\frac{S}{2})^p = 2^{1-p} S^p
\end{align*}

Hence we attain the maximum value of $||\mathbf{x}||_p^p$ when $\mathbf{x}=\mathbf{v}$. We will now show that $\mathbf{v}$ is indeed the maximum.

First we note that since $\sum_{i=1}^n x_i=0$, the sum of all the negative $x_i$'s must be exactly the opposite of the sum of all the positive $x_i$'s. Furthermore, we can partition the $x_i$'s into two sets, P and N, where
\begin{align*}
    N:&=\{x_i | x_i < 0, x_i \in \mathbf{x}\}, \text{and } \sum_{x_i<0} x_i=-\frac{S}{2} \\
    P:&=\{x_i | x_i \geq 0, x_i \in \mathbf{x}\}, \text{and } \sum_{x_i\geq 0} x_i=\frac{S}{2}
\end{align*}
If we have exactly one non-zero value in both P and N, then our vector must be $\mathbf{v}$. W.L.O.G., assume we have two non-zero values, $x_j\geq x_k>0$ and $x_j,x_k\in P$.




\textit{Claim:} $(x_j + x_k)^p > x_j^p + x_k^p$.

Let's consider the $L^p$-$space$ on $\mathbb{R}^2$ with $p$-$norm$ $||\mathbf{u}||_p := (|u_1|^p+|u_2|^p)^{1/p}$. Let $\mathbf{y}=(x_j,0)$ and $\mathbf{z}=(0,x_k)$. Applying \textit{Minkowski Inequality} gives us $x_j + x_k > (x_j^p + x_k^p)^{1/p}$ and the claim holds.

Following what we have shown above, $||\mathbf{x}||_p^p$ is strictly increasing if we replace $x_j$ and $x_k$ with $x_j*=0$ and $x_k*=x_j+x_k$. We note that this replacement does not change the mean or the value of $S$. By symmetry, the same holds for $N$. We may continue this replacement process until there's only one non-zero value left in both $N$ and $P$, and since this process monotonically increases $||\mathbf{x}||_p^p$, we conclude that $||\mathbf{x}||_p^p \leq 2^{1-p} S^p$ and we attain the maximum when $\mathbf{x}=\mathbf{v}$. We will now use the above lemma to prove a useful theorem.

\textbf{Theorem 1.} (Optimal Low Precision Pre-normalizer Theorem). \textit{Let $\mathbf{x}=\{x_1, x_2, ..., x_n\}$ be a finite vector of real numbers with $\sum_{i=1}^n x_i=0$. Let $M$ be the maximum value of our low precision path. Then,}
\begin{align*}
    f(\mathbf{x}) = \frac{\mathbf{x}}{\frac{1}{2}(\frac{2}{M})^{1/p}\sum_{i=1}^n |x_i|}
\end{align*}
\textit{is an optimal $L_p$-$norm$ pre-normalizer for this compute path.}

\textit{Proof:} From Lemma 1, we know that $||\mathbf{x}||_p^p$ attains the maximum value when $\mathbf{x}=\mathbf{v} = \{-\frac{S}{2}, 0, ..., 0, \frac{S}{2}\}$, where $S$ is the $L_1$-$norm$ of $\mathbf{x}$. Thus it suffices to prove that $f(\mathbf{v})$ satisfies \textit{Definition 2}.

\begin{align}
    ||f(\mathbf{v})||_p^p & = \sum_{j=1}^n |\frac{v_j}{\frac{1}{2}(\frac{2}{M})^{1/p}\sum_{i=1}^n |v_i|}|^p \\
    & = \Big(\frac{|-\frac{S}{2}|}{\frac{1}{2}(\frac{2}{M})^{1/p}\sum_{i=1}^n |v_i|}\Big)^p + \\
    & \textcolor{white}{hello} \Big(\frac{|\frac{S}{2}|}{\frac{1}{2}(\frac{2}{M})^{1/p}\sum_{i=1}^n |v_i|}\Big)^p \\
    & = \Big(\frac{\frac{S}{2}}{\frac{1}{2}(\frac{2}{M})^{1/p} S}\Big)^p + \Big(\frac{\frac{S}{2}}{\frac{1}{2}(\frac{2}{M})^{1/p} S}\Big)^p \\
    & = \frac{M}{2} + \frac{M}{2} = M
\end{align}

As shown above, the largest possible value attainable after applying our pre-normalizer is precisely $M$, the maximum value of our low precision path. \hfill $\square$

\textbf{Corollary 1.} \textit{$f(\mathbf{x}) = \frac{\mathbf{x}}{\frac{\sqrt{2}}{512}\sum_{i=1}^n |x_i|}$ is an optimal low precision pre-normalizer for $L_2$-$norm$ on the FP16 compute path.}

On a practical note, the pre-normalizer we used for our experiment was the one from Lemmas A1 and A2 (\ref{sec:appendix}) with $n=512$, which gave a slightly lower normalization constant than what Corollary 1 suggests. This worked well in our setup because attaining or even getting close to the maximum value as stated in Lemma 1 requires atypical distribution of values with very few extreme values and everything else being 0. This does not happen in practice, however, with the most common distribution of values observed being Gaussian.

\subsection{Scaling of Softmax}
\label{sec:softmax}

Another common constraint on hardware accelerators is their limited support in complex operations. For example, hardware accelerators may choose to omit support for exponential operations \cite{hu2018efficient, li2018efficient}. In such cases, we seek to implement such operations in memory instead, namely using lookup tables (LUT). However, since LUTs are slow and expensive in terms of memory consumption, we would like the tables to be as small as possible. To this end, we introduce a technique called conditional re-scaling for softmax layers:

\begin{align*}
    \mathbf{x}=
    \begin{cases}
    \frac{4096\mathbf{x}}{\max(\mathbf{x})} & \text{if $max(\mathbf{x})>4096$} \\
    \mathbf{x} & \text{otherwise.}
    \end{cases}
\end{align*}

To interpret the above transformation, we first assume that our LUT gives reasonably accurate approximation for $x_i$'s below 4096. Next we take FP16 as an example of our low precision compute paths. We note that for values greater than 4096, gaps between values jump in increments of 4 according to IEEE 754-2008 \cite{4610935}. Under such scenario, the softmax function behaves similarly to an argmax operation. Since gaps of values between 2048 and 4096 jump in increments of 2, the ``argmax behavior" is largely preserved after the re-scaling and exponentiation. 

\begin{figure}[h]
    \centering
    \includegraphics[width=0.47\textwidth]{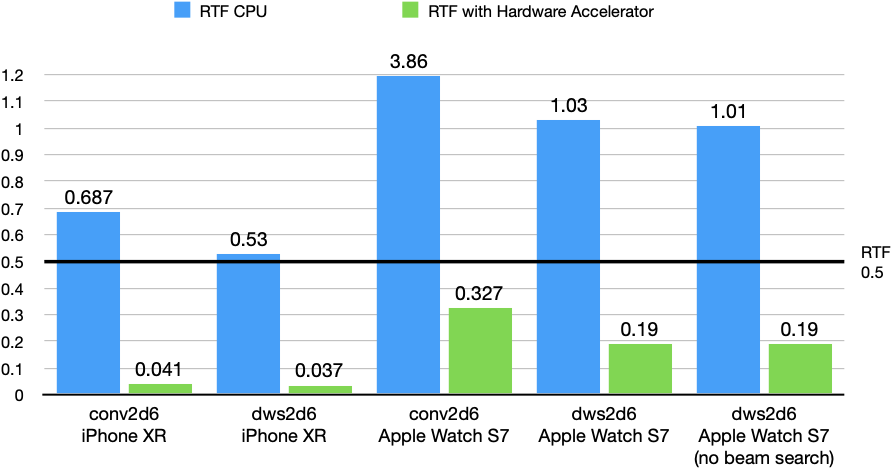}
    \caption{Realtime Factor (RTF) of the original Conformer CTC vs Depthwise Separable Convolution (DWS) architectures. Blue and green bars represent the RTF on CPU and hardware accelerators, respectively. We also added a horizontal line at 0.5 to illustrate required RTF for ASR to process in realtime.}
    \label{fig:paper-rtf}
\end{figure}

\begin{figure}[t]
    \centering
    \includegraphics[width=0.47\textwidth]{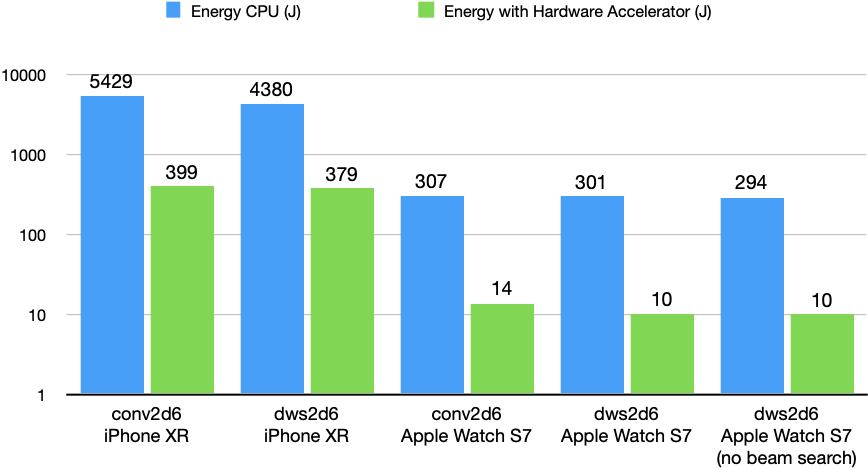}
    \caption{Energy consumption (in joules) for 200 queries of the original Conformer CTC vs Depthwise Separable Convolution (DWS) architectures. Blue and green bars represent the values on CPU and hardware accelerators, respectively. The y-axis is in log scale.}
    \label{fig:paper-energy}
\end{figure}

\section{Experiments and Results}
\label{sec:expriment}

\subsection{Setup}
\label{sec:setup}

The training corpus contains 17k-hour audio-transcript pairs where the audio is randomly sampled from anonymized virtual assistant queries and human-annotated. 
We curate 20k queries in the same manner to form an accuracy test set. 
We use it to examine the accuracy of the optimizations.
200 queries are sampled from the accuracy test set and serve as the performance test set. 
The audio is decoded lightweightedly with CTC prefix beam search so as to rule out as many computationally intensive components as possible \cite{DBLP:conf/icml/GravesFGS06}.
The data choice and the training recipe do not play important role in the experiments because the proposed methods focus on hardware acceleration.
The experiments are conducted on iPhone XR and Apple Watch Series 7. 

Two models (\textit{conv2d6} and \textit{dws2d6}) are trained with the same hyper-parameters but minor difference in subsampling strategy, summarized in 
Appendix \ref{sec:hypparam}.
Another two models (\textit{conv2d6x22} 
and \textit{dws2d6x22}) are trained with the same configuration except that the input to the first Conformer block is scaled by a factor of square root of the IO dimension described in \cite{DBLP:conf/nips/VaswaniSPUJGKP17}. 
Additionally we decode greedily on watch to show that \textbf{encoder's workload dominates}.

\subsection{Performance}

High performance is critical in an ASR system in order to process a user's request in real time. To benchmark the performance, we define a notion of Realtime Factor (RTF) as $RTF=processingTime/audioDuration$. It is clear from the definition that lower RTF values are desirable. On real devices, users may often multitask or the operating system may occasionally use computing resources in the background. Therefore an RTF value of at least 0.5 is 
a reasonable target.
As we can see from Figure \ref{fig:paper-rtf}, models running on CPUs do not meet our RTF target of 0.5 and the performance is substandard on the watch. By leveraging deep learning hardware accelerators, we are able to bring the RTF down by an order of a magnitude for both model variants and achieve the performance goal.
On Apple Watch, it is 5.26 times faster.

\begin{table}
    \centering
    \begin{tabular}{l|c|l|c}
        \hline
        \textbf{model w/} & \multirow{2}{*}{\textbf{overflow}} & \textbf{model w/o} & \multirow{2}{*}{\textbf{overflow}} \\
        \textbf{multiplier} & & \textbf{multiplier} &  \\
        \hline
         conv2d6x22 & 6.85\% & conv2d6 & 3.26\% \\
         dws2d6x22 & 6.85\% & dws2d6 & 0.25\% \\
         \hline
    \end{tabular}
    \caption{Layernorm overflow statistics when the proposed transform in Section \ref{sec:Layernorm} is not applied}
    \label{tab:ln-over}
\end{table}

\subsection{Energy}

Another important aspect to consider when executing an ASR system on device is the energy consumption. Energy consumption is particularly vital on mobile devices and wearables. We report the energy reduction from using hardware accelerators in Figure \ref{fig:paper-energy}, where we again see reduction by an order of a magnitude. 

\subsection{Numeric Stability}

\begin{figure}
    \centering
    \includegraphics[width=0.475\textwidth]{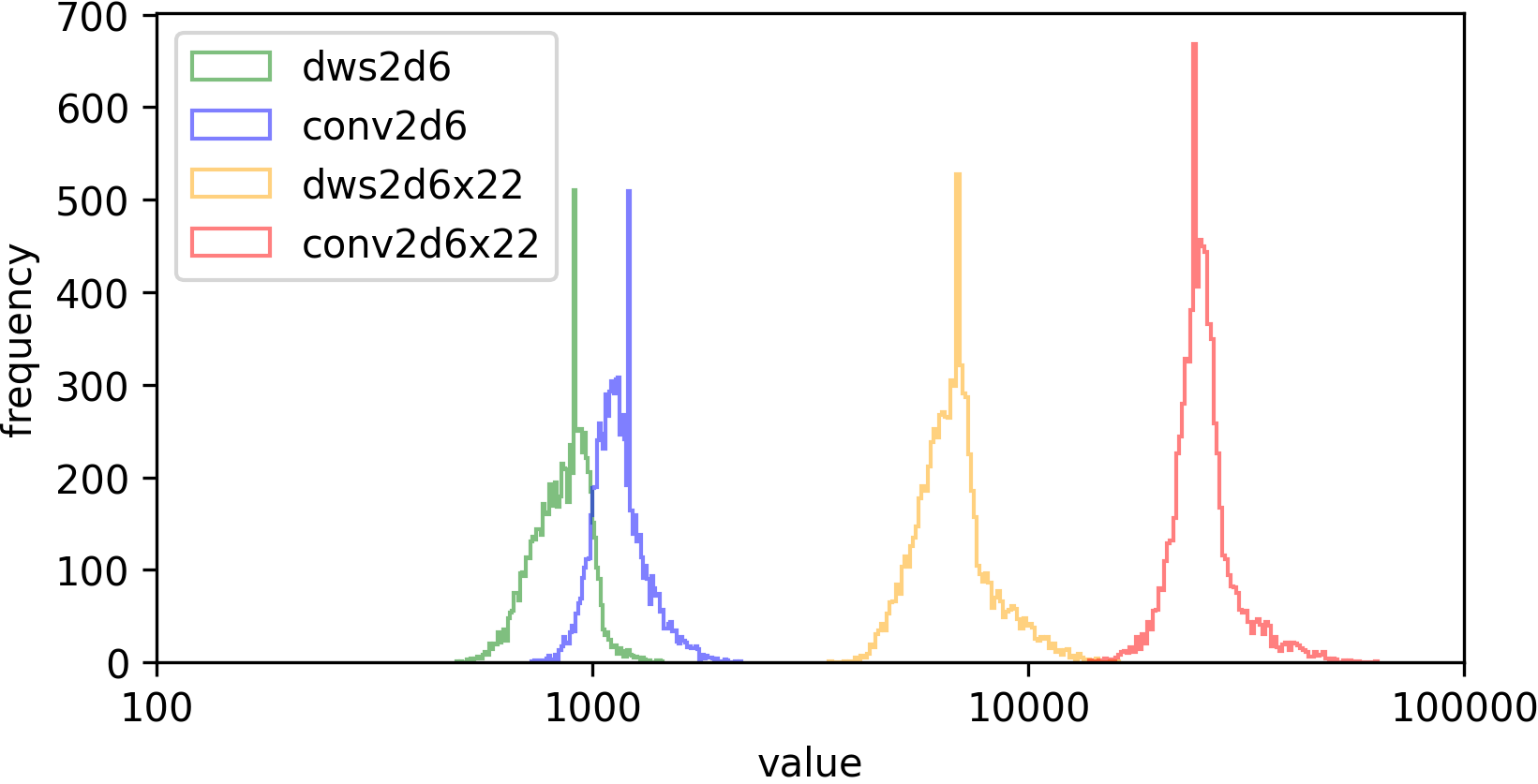}
    \caption{Distribution of the max value between vanilla convolution and DWS in log scale.}
    \label{fig:subsampling}
\end{figure}

\begin{figure}
    \centering
    \includegraphics[width=0.475\textwidth]{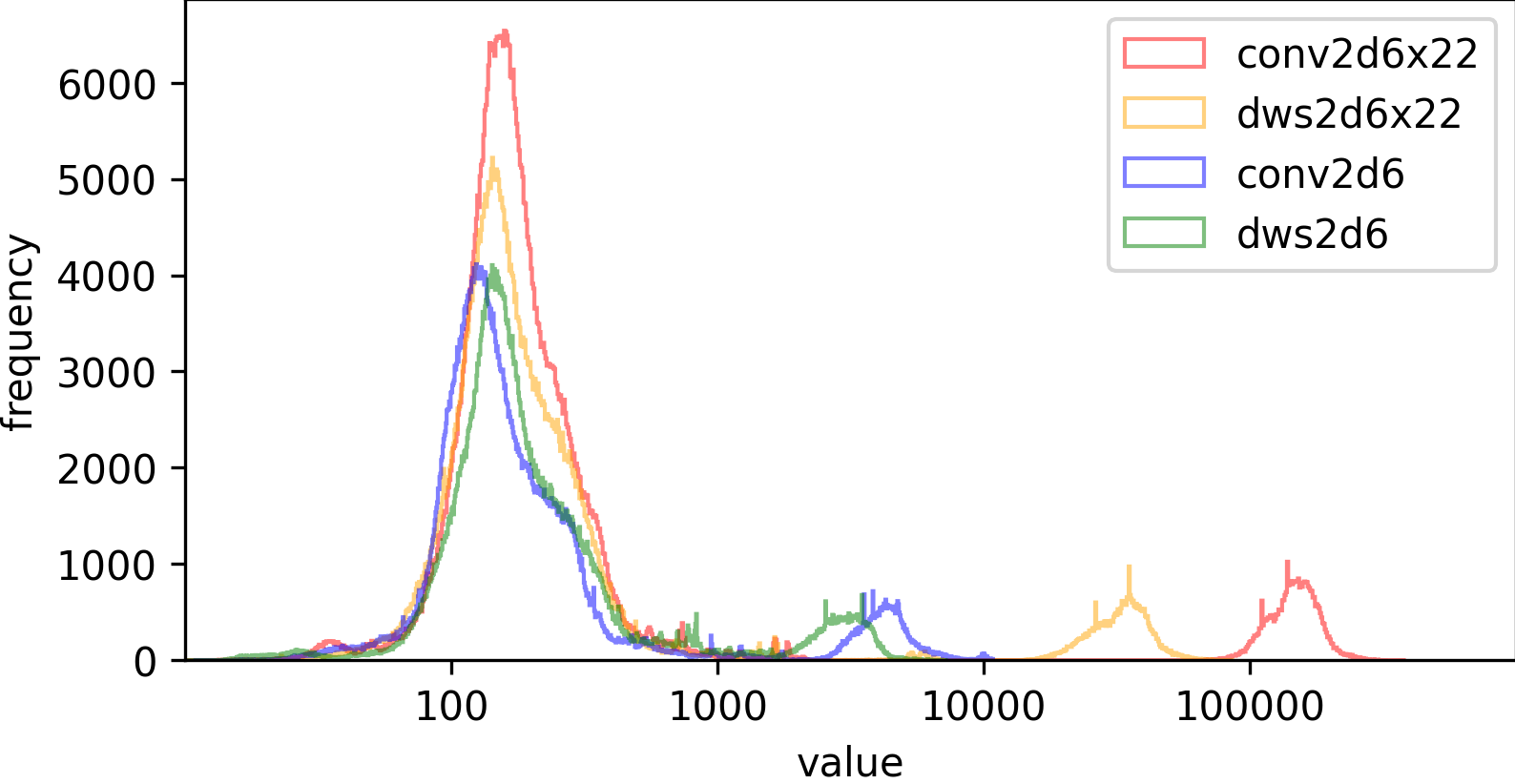}
    \caption {Distribution of Layernorm's input's max value in log scale.}
    \label{fig:ln-input}
\end{figure}

In Figure \ref{fig:subsampling} we compare the distribution of maximum value of each chunk's subsampling output during a chunk-based decoding procedure between vanilla convolution and DWS over the performance test set.
Empirically the dynamic range of DWS subsampling is a few times smaller than that of the vanilla 2D convolution. 
When we compare \textit{dws2d6} against \textit{dws2d6x22} or \textit{conv2d6} against \textit{conv2d6x22}, we observe one or two orders of magnitude dynamic range increase introduced by the square root multiplier. 
Therefore, switching to DWS and removing the multiplier are crucial to keep the subsampling in low-precision-friendly area.
Similarly, we plot the distribution of maximum value of each chunk for the Layernorms in Figure \ref{fig:ln-input}. 
Due to residual connections, the enlarged effect of the subsampling output is cascading, i.e. large subsampling output increases the chance of overflow in upper layers. In Table \ref{tab:ln-over}, we collected overflow statistics of the un-modified Layernorm.

\begin{table}
    \centering
    \begin{tabular}{p{0.16\textwidth}|p{0.12\textwidth}p{0.12\textwidth}}
        \hline
        \textbf{model} & \textbf{WER} (FP16) & \textbf{WER} (FP32) \\
        \hline
        conv2d6 & 4.45\% & 4.41\% \\
        dws2d6 & 4.55\% & 4.56\% \\
        conv2d6x22 & 4.57\% & 4.47\% \\
        dws2d6x22 & 4.57\% & 4.49\% \\
        \hline
        conv2d6x22 & \multirow{2}{0.11\textwidth}{4.76\%}  & \multirow{2}{0.11\textwidth}{4.72\%} \\
        + \small{modified Softmax} & \\
        \hline
    \end{tabular}
    \caption{WER comparison of FP16 and FP32}
    \label{tab:wer}
\end{table}

\subsection{Quality}

We compare the WER of the models on various settings and observed that
(1) The difference between FP16 and FP32 is negligible, (2) DWS and vanilla convolution yield almost same accuracy and (3) feature scale-up from the transformer work is not necessary.
\textit{conv2dx22} has an almost overflow dynamic range. 
We apply the softmax modification in Section \ref{sec:softmax} on top of \textit{conv2dx22}. 
There is a slight WER regression. However, such WER regression does not affect user experience when WER is already low. 

\section{Conclusions}
\label{sec:conclusion}

Through architectural and numerical optimizations, we demonstrate that Conformer CTC ASR models are capable of running on resource-constrained devices such as mobile phones, and wearables. 
The optimizations preserve recognition accuracy while performing faster than real time and consuming lesser energy.
Our theoretical findings of techniques in numerical stabilization is applicable to a wide range of deep learning models and computing tasks.

\bibliography{anthology,custom}

\clearpage

\appendix

\section{Hyper Parameters}
\label{sec:hypparam}

\begin{description} 
    \item[conv2d6x22] follows the recipe of \cite{DBLP:conf/interspeech/YaoWWZYYPCXL21, DBLP:conf/interspeech/ZhangWPSY00YP022}, where the subsampling output is multiplied by $\sqrt{512}$ before being fed into the first conformer layer. The multiplier is originated from the transformer work \cite{DBLP:conf/nips/VaswaniSPUJGKP17}.
 Its hyper-parameters are summarized in Table \ref{tab:hyperparam}. 
    \item[dws2d6x22] is produced by replacing vanilla convolutional subsampling with depthwise separable convolution (DWS). Their difference is compared in Table \ref{tab:subsampling}.
    \item[conv2d6] is indentical to conv2dx22 except that multiplier is not applied.
    \item[dws2d6] is same as dws2dx22 but without applying the multiplier.
\end{description}

\begin{table}[h]
    \centering
    \begin{tabular}{ll}
        \hline
        \textbf{hyper-parameters} & \textbf{values} \\
        \hline
        \#layers (encoder) & 12  \\
        \#layers (decoder) & 3 \\
        \#heads & 8 \\
        layer IO dimension & 512 \\
        feedforward dimension & 2048 \\
        \hline
    \end{tabular}
    \caption{Common hyper-parameters in the experiments}
    \label{tab:hyperparam}
\end{table}

\begin{table}[h]
    \centering
    \begin{tabular}{l|llll}
        \hline
        \textbf{model} & \textbf{channel} & \textbf{kernel} & \textbf{stride} & \textbf{group} \\
        \hline
        \multirow{2}{*}{conv2d6} & 1 $\rightarrow$ 512 & (3,3) & (2,2) & 1 \\
                & 512 $\rightarrow$ 512 & (5,5) & (3,3) & 1 \\
        \hline
        \multirow{3}{*}{dws2d6} & 1 $\rightarrow$ 512 & (3,3) & (2,2) & 1 \\
               & 512 $\rightarrow$ 512 & (5,5) & (3,3) & 512 \\
               & 512 $\rightarrow$ 512 & (1,1) & (1,1) & 1 \\
        \hline
    \end{tabular}
    \caption{Different subsampling hyper-parameters. Convolution in the same group are applied sequentially.}
    \label{tab:subsampling}
\end{table}

\section{Mean Absolute Deviation Normalization on Example Distributions}
\label{sec:appendix}

\textbf{Definition A1.} \textit{A desirable low precision pre-normalizer maps a distribution of values to a bounded region, $[-C, C]$, for some small $C$.}

\textbf{Lemma A1.} \textit{$f(\textbf{x}) = \frac{\textbf{x}}{\frac{1}{n}\sum_{i=1}^{n}|x_i|}$ is a \textit{desirable} low precision pre-normalizer for uniform distributions.}

\textit{Proof}: suppose $X\sim unif[-L,L]$ and $\textbf{x}$ is a vector of $x_i$'s sampled from $X$. Consider the limit of the denominator of our normalizer as $n\to\infty$,

\begin{align*}
    \lim_{n\to\infty} \frac{1}{n}\sum_{i=0}^n |x_i| & = \mathbb{E} [|\textbf{x}|] = \int_{-L}^{L} \frac{|x|}{2L} dx = \frac{L}{2}.
\end{align*}

Thus, $f(\textbf{x})=\frac{2\textbf{x}}{L}\sim unif[-2,2]$.

\textbf{Lemma A2.} \textit{$f(\textbf{x}) = \frac{\textbf{x}}{\frac{1}{n}\sum_{i=1}^{n}|x_i|}$ is a \textit{desirable} low precision pre-normalizer for normal distributions.}

\textit{Proof}: suppose $X\sim N(0,\sigma)$ and $\textbf{x}$ is a vector of $x_i$'s sampled from $X$. Consider the limit of the denominator of our normalizer and $n\to\infty$,

\begin{align*}
    \lim_{n\to\infty} \frac{1}{n}\sum_{i=0}^n |x_i| & = \mathbb{E} [|\textbf{x}|] \\
    & = \frac{1}{\sigma \sqrt{2\pi}} \int_{-\infty}^{\infty} |x|e^{-\frac{1}{2}(\frac{x}{\sigma})^2} dx \\
    & = \frac{2}{\sigma \sqrt{2\pi}} \int_{0}^{\infty} xe^{-\frac{1}{2}(\frac{x}{\sigma})^2} dx \\
    & \textcolor{white}{zero} (by\ symmetry) \\
    & = \sqrt{\frac{2}{\pi}}\sigma.
\end{align*}

Let $x=k\sigma$ for some real $k$, $f(x)=k\sqrt{\frac{\pi}{2}}$. When $k=\pm 4$, $f(x)=\pm 5.01$. In other words, $f(x)\in [-5.01,5.01]$ with $99.99\%$ probability.

The two lemmas above illustrate the effect of our MAD normalizer on a couple of common distributions. Empirically, we observed no overflow during our subsequent Layernorm computation after we prepended our pre-normalizer. Let us now look at the theory behind a bit more rigorously.

\end{document}